\documentclass[sigconf]{acmart}
\renewcommand\footnotetextcopyrightpermission[1]{} 
\usepackage[utf8]{inputenc}
\usepackage[super]{nth}
\usepackage{booktabs}
\usepackage{url}
\usepackage{color}
\usepackage{makecell}

\usepackage{multirow}
\usepackage{graphicx}
\usepackage{subcaption}
\usepackage{hyperref}
\usepackage{pifont}
\usepackage{algorithm}
\usepackage{algorithmic}
\newcommand{\xmark}{\ding{55}}%
\usepackage{amsmath}
\usepackage{arydshln}
\usepackage[font=small,labelfont=bf]{caption}
\usepackage{diagbox}

\begin{document}
\title{Multi-characteristic Subject Selection from Biased Datasets} 

\author{Tahereh Arabghalizi}
\affiliation{%
  \department{Department of Computer Science} 
  \department{School of Computing and Information}
   \institution{University of Pittsburgh}
  \streetaddress{210 S Bouquet St}
  \postcode{15213}}
\email{tahereh.arabghalizi@pitt.edu}

\author{Alexandros Labrinidis}
\affiliation{%
  \department{Department of Computer Science} 
  \department{School of Computing and Information}
   \institution{University of Pittsburgh}
  \streetaddress{210 S Bouquet St}
  \postcode{15213}}
\email{labrinid@cs.pitt.edu}

\begin{abstract} 
Subject selection plays a critical role in experimental studies, especially ones with human subjects. Anecdotal evidence suggests that many such studies, done at or near university campus settings, suffer from \emph{selection bias}, i.e., the too-many-college-kids-as-subjects problem. Unfortunately, traditional sampling techniques, when applied over biased data, will typically return biased results. In this paper, we tackle the problem of multi-characteristic subject selection from biased datasets. We present a constrained optimization-based method that finds the best possible sampling fractions for the different population subgroups, based on the desired sampling fractions provided by the researcher running the subject selection. We perform an extensive experimental study, using a variety of real datasets. Our results show that our proposed method outperforms the baselines for all problem variations by up to 90\%.
\end{abstract}

%

%

\maketitle

\section{Introduction}
Studies involving human subjects are becoming increasingly common, and are often a required element in many types of research, from medical to social and everything in between. Best practices in designing and running such studies dictate that subject selection must be fair (in terms of sharing the benefits and burdens of the research), and researchers should not exclude subjects based on characteristics such as age, gender, race, religion, sexual orientation, etc.\cite{miser2005educational,pech2007understanding}. 

The recent \emph{"Black Lives Matter"} movement has demonstrated vividly the difference between \emph{not being racist} and \emph{being anti-racist}. In other words, the need to go from not excluding people deliberately (e.g., based on race) to being \emph{actively inclusive}. One early example of unintended consequences when inclusivity is not a design principle in a data-driven project is the Boston pothole detection mobile app.  The app automatically identified potholes by detecting the bumps of cars driving over potholes (while the app was running on a smartphone in the car). The resulting maps showed entire neighborhoods virtually pothole-free; unsurprisingly, these were the more impoverished neighborhoods, where not enough people had smartphones at the time~\cite{harvardbigdata}.

The driving principle of this paper is \emph{how to guarantee inclusiveness for subject selection in user studies}. Traditionally, subject selection involves randomized sampling techniques. However, such techniques assume that it is perfectly acceptable for the characteristics of the selected population to align with the ones of the "input" population. For small-scale studies, that is resolved by manual intervention by the researchers running the study. For larger-scale studies, this is often unresolved and leads to instances of the too-many-college-kids-as-subjects problem, where studies performed at or near university campus settings rely primarily on subjects not representative of the broader population. This phenomenon, which is due to misrepresentation in data, is known as \emph{Selection Bias} and can lead to unreliable analyses and unfair decisions~\cite{ntoutsi2020bias,cortes2008sample}.

\smallskip
\noindent\textbf{Motivating Example \#1:}
\begin{figure}[t!] 
\centering
  \begin{subfigure}[b]{0.49\linewidth}
    \centering
    \includegraphics[width=1.1\linewidth]{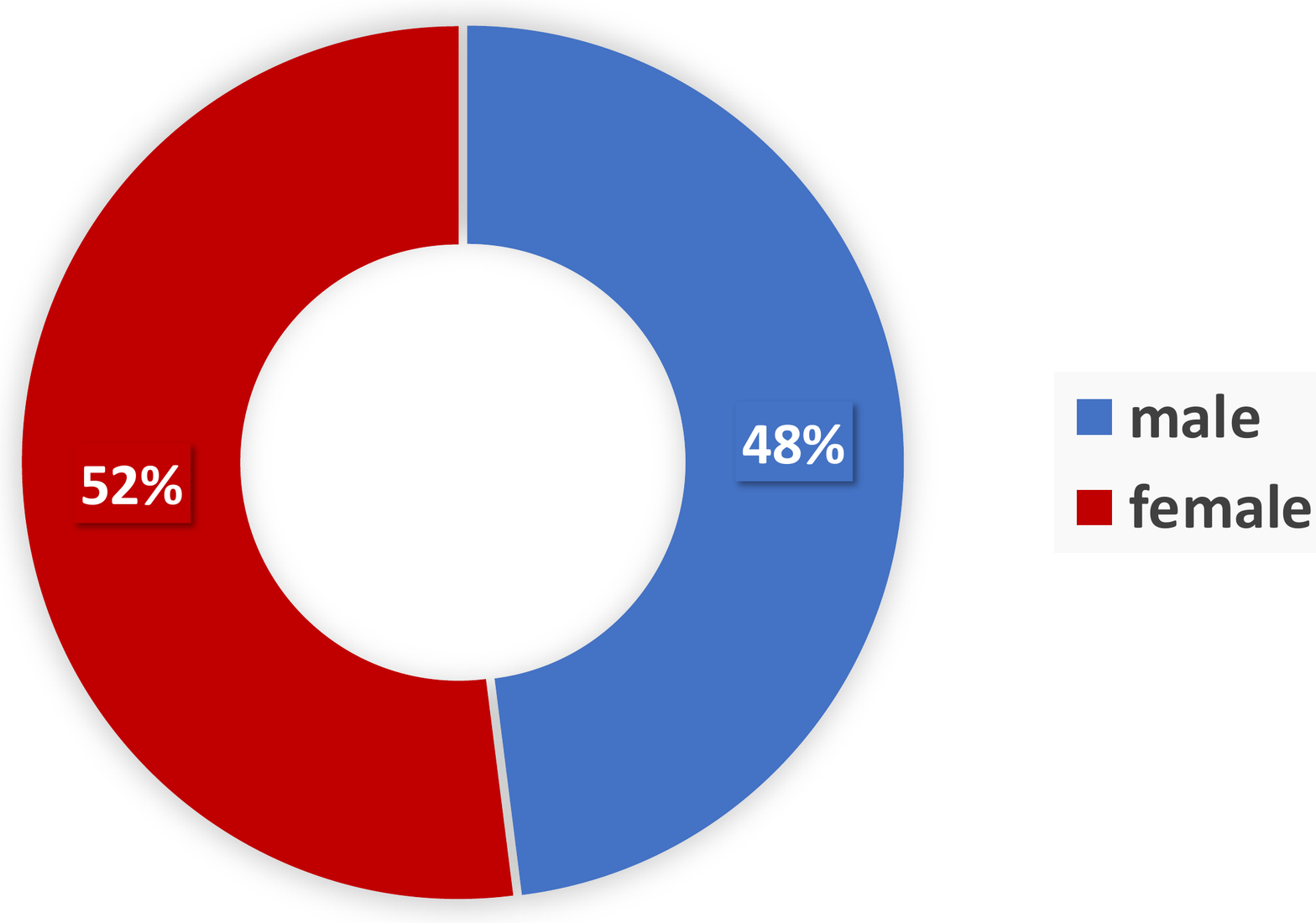}
    \caption{gender - Pittsburgh census} 
    \label{fig:pitt_age} 
    \vspace{0.5ex}
  \end{subfigure} 
  \begin{subfigure}[b]{0.49\linewidth}
    \centering
    \includegraphics[width=1.1\linewidth]{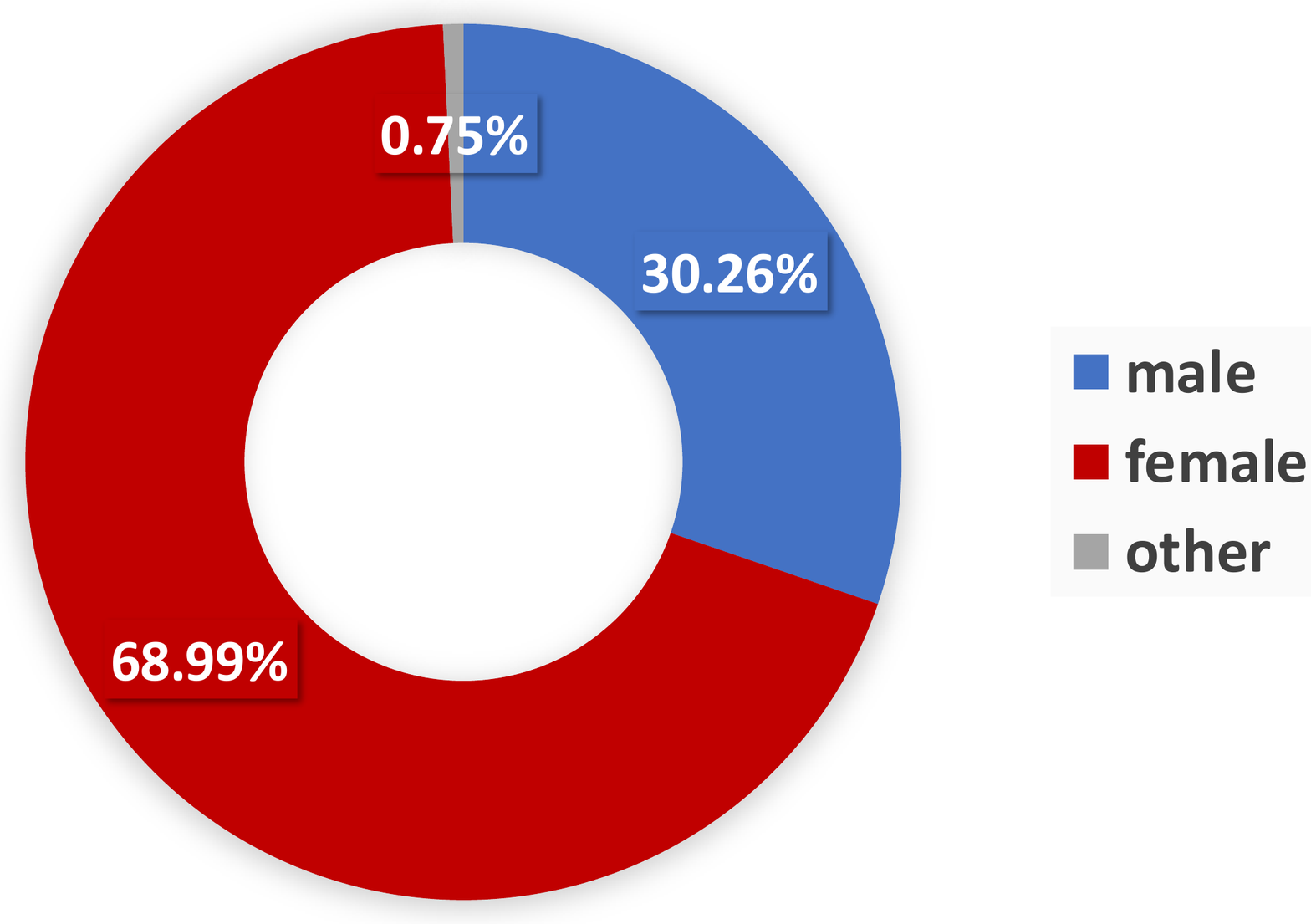}
    \caption{gender - our survey} 
    \label{fig:pitt_gender} 
    \vspace{0.5ex}
 \end{subfigure}
 
\caption{gender distribution - Pittsburgh census vs our survey}
\label{fig:motiv1}   
\end{figure}  
Let us assume we want to select subjects for a field experiment to study the impact of incentives in changing people's public transportation habits. This example is the actual motivation behind this work and a need we have in the PittSmartLiving project (\url{https://PittSmartLiving.org}), whose goal is to encourage pro-social transportation behavior by providing real-time information and incentives to bus riders. For example, a field experiment subject will see a notification that says: \emph{"the bus you are waiting for will be full, if you take the following one, the coffee shop around the corner will give you a \$2 discount on coffee"}. Obviously, for such a study to be impactful, we need to make sure we have a representative sample population when compared to that of the entire city.

\begin{figure}[ht!] 
\centering
 \begin{subfigure}[b]{1\linewidth}
    \centering
    \includegraphics[height=4.2cm, width=0.90\linewidth]{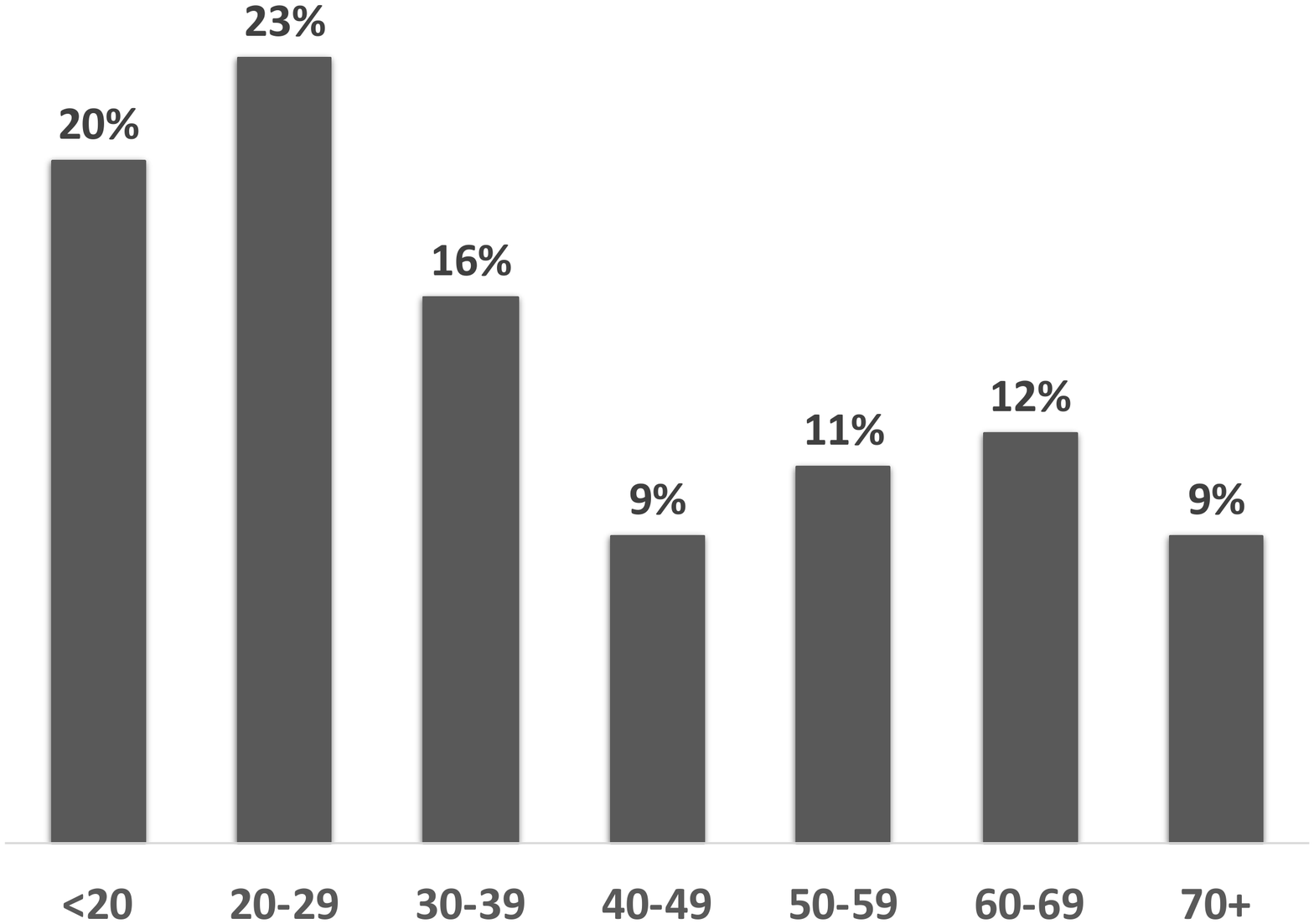}
    \caption{age distribution - Pittsburgh census} 
    \label{fig:pitt_age} 
    \vspace{0.5ex}
  \end{subfigure} 
  \begin{subfigure}[b]{1\linewidth}
    \centering
    \includegraphics[height=4.2cm, width=0.90\linewidth]{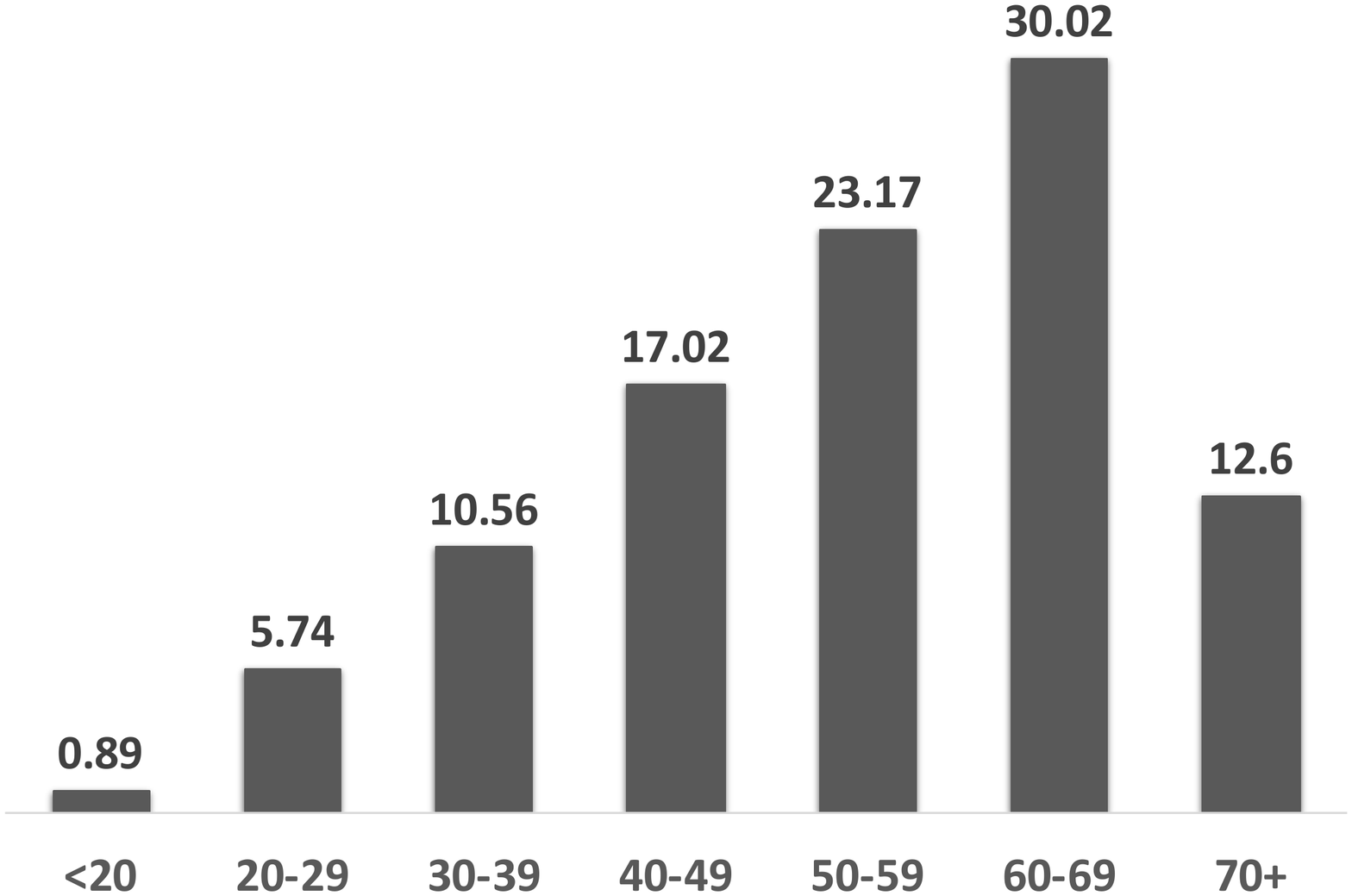}
    \caption{age distribution - our survey} 
    \label{fig:pitt_gender} 
    \vspace{0.5ex}
 \end{subfigure}

\caption{age distribution - Pittsburgh census vs our survey}
\label{fig:motiv2}   
\end{figure}

When we compare the characteristics of the participants of a prior transportation survey we performed to that of the population of Pittsburgh (as summarized in the census), we saw a big discrepancy. 
The two pie charts in Figure \ref{fig:motiv1} demonstrate the breakdown by gender in the Pittsburgh census\cite{censusreporter} vs. in our survey (Section \ref{lab:data}). While, in the census data, the percentage of female subjects accounted for over half the population, in the survey data, the number of females is more than twice the number of males, and there is also a tiny proportion (0.75\%) listed as "other gender." Furthermore, bar charts in Figure \ref{fig:motiv2} compare the age distribution in the Pittsburgh census vs. our survey data. The survey data is entirely biased and not representative of the whole population. 
So if we are to create a sample out of the people surveyed that would be representative of the Pittsburgh population, \emph{we would need to match the distribution of characteristics of the selected subjects to that of the general population. } And this is the exact point where traditional sampling-based subject selection techniques would fail, and we need new methods. 

\smallskip
\noindent\textbf{Motivating Example \#2:}
As a second motivating example, consider a human subjects study for Type II Diabetes. The researchers running the study want to include people who have the disease and create a cohort with a balanced mix of male, female, plus people that do not identify as male or female, i.e., "other gender." Additionally, they would like to primarily have people in their 30s and 40s participate in the study, plus a few in their 20s and 50s.  
If we try to cast the above requirements into a set of desired/ideal distributions for the different characteristics, we run into the problem that it is difficult to determine specific percentages (like one would get from the census). For example, how much should the researchers allocate to the different age groups? 35\% to each of the 30s and 40s vs. 15\% to the 20s and 50s? How about 40\% to the first group and only 10\% to the second one? A fitting solution here should support \textbf{defining ranges of ideal percentages} so that the researchers, for example, can specify 30-40\% for the first group and 10-20\% for the second group. And such ranges could start at 0\%, to account, for example, for the possibility of not having a subject identifying as "other gender."

Another interesting desideratum is the ability to specify generalized constraints on the subject characteristics. For example, for our fictional diabetes study, the researchers may also want to make sure all neighborhoods from a city are equally included in the subject pool. Although this constraint can be trivially specified with a function, it cannot be part of any known sampling technique.  
 
\smallskip
Traditional subject selection methods such as Stratified Random Sampling either do not support the situations from our two motivating examples or provide expensive solutions, especially when dealing with multiple data characteristics~\cite{taherdoost2016sampling}. There are some biased sampling techniques such as Disproportionate Stratified Random Sampling by which some of the groups will be over-represented, and some will be under-represented~\cite{haque2010sampling}. However, these will not work for our second motivating example.

\smallskip
\noindent \textbf{Contributions:} 
In this work, we propose a novel approach that not only resolves the problem of multi-characteristic subject selection from biased datasets but also supports different variations of the problem, including range-based and constraint-based (generalized). To the best of our knowledge, our work is unique, and no work has ever addressed the same problem or proposed a solution for it. 
In particular, we make the following contributions:
\begin{enumerate}
    \item We formulate the problem of multi-characteristic subject selection from biased datasets in the form of an optimization problem that minimizes the distance between the ideal and the final sampling distributions. Our proposed method works with different variations of the problem, including fixed, range-based, and generalized (with more constraints and objective functions).
    \item We propose an algorithm that computes the joint ideal percentage in multi-characteristic problems.
    \item We perform several experimental evaluations using data from real-world applications and compare our proposed method to three different baselines. Our results show that our proposed method outperforms the baselines by up to 90\% for some experiments.
\end{enumerate}
\section{Problem Statement}
The main objective of this work is to develop a subject selection method that allocates the best possible sampling percentage to each subgroup of the population who shares similar characteristics. What we mean by "the best possible sampling percentage" is that the percentage of \emph{final selection} in each subgroup should be as close as possible to the percentage of \emph{ideal selection} in the same subgroup while satisfying a set of given constraints. In this work, we assume that the researcher knows the total sample size or has already obtained it based on the required precision, confidence levels, and population variability.

The first step before subject selection is dividing the population into homogeneous subgroups called \emph{"strata"} which are collectively exhaustive and mutually exclusive\cite{wiki2}. 
For example, one stratum could be (male, 30-39) and another stratum could be (female, 60-69). Since we want to include several data characteristics in the subject selection problem, the ideal percentage given for characteristic groups (e.g. 40\% male and 60\% female) cannot be used for the strata. Therefore, we propose an algorithm (see section \ref{lable:jointideal}) that calculates the \emph{"joint ideal percentage"} for each stratum using the provided ideal percentages for the characteristic groups that are involved in that stratum. 

We address three variations of the stated problem as follows:\\
\begin{itemize}
    \item \textbf{Fixed} 
(e.g., Motivating Example \#1) In this variation, we are provided with a fixed ideal percentage for each group of data characteristics. For example, male: 48\%\ and female: 52\%\ . \\
    \item \textbf{Range-based} 
(e.g., Motivating Example \#2) In this variation, some or all of the ideal percentages for characteristic groups are range-based. For instance, the percentage of people between 18-21 years old can be between 10\%\ and 20\%. \\
    \item \textbf{Generalized} 
(e.g., Motivating Example \#2) This variation is, in fact, an extension of the range-based variation where a set of constraints or objective functions, on the characteristic groups, need to be satisfied. For example, the number of female subjects must be twice as many as the number of male subjects or the payments made to subjects must be evenly distributed across all states.
\end{itemize}

\section{Baseline Algorithms}
In this section, we present the evaluation metric, our proposed algorithm for computing the joint ideal percentage, the variations of the baselines and the baseline methods.
\subsection{Evaluation Metric}
As mentioned earlier, the purpose of this work is to allocate a percentage to each stratum such that the distance between the joint ideal selection/distribution and the allocated final selection/distribution becomes as small as possible. We considered different distance functions, which included \emph{Cosine distance}, \emph{Euclidean distance}, and \emph{Kullback–Leibler divergence}\cite{kldivergence}. We settled on using the \emph{Cosine Distance} as our evaluation metric to measure the distance between these two vectors in a multi-dimensional space. The Cosine Distance is easier and faster to implement; it is also easier to understand compared to the other metrics while their results are quite similar. 

The Cosine Distance (equation \ref{eq:distance}) can be computed using Cosine Similarity (equation \ref{eq:cosine}) that captures the cosine of the angle between the two vectors\cite{wiki1}.
The space dimension D (or the number of strata) for these two vectors can be computed by equation \ref{eq:dimension}. 
\begin{equation} 
\small
\label{eq:distance} 
Cosine Distance = 1 - Cosine Sim(\overrightarrow{V_{F}},\overrightarrow{V_{JI}}) 
\end{equation} 
\begin{equation} \small \label{eq:cosine}Cosine Sim(\overrightarrow{V_{F}},\overrightarrow{V_{JI}}) = {\overrightarrow{{V_{F}}} \overrightarrow{{V_{JI}}} \over \|\overrightarrow{{V_{F}}}\| \|\overrightarrow{{V_{JI}}}\|} = \frac{ \sum_{h=1}^{D}{{F_h}*{JI_h}}}{ \sqrt{\sum_{h=1}^{D}{({F_h})^2}}*\sqrt{\sum_{h=1}^{D}{({JI_h})^2}} }\end{equation} 
where $\overrightarrow{V_{F}} = (F_{1}, F_{2}, ...\;, F_{D})$ and $\overrightarrow{V_{JI}} = (JI_{1}, JI_{2}, ...\;, JI_{D})$ are the final and joint ideal percentage vectors respectively. The components of vector $\overrightarrow{V_{JI}}$ are joint ideal percentages for the strata that come from algorithm \ref{alg:jointideal}.
\begin{equation} 
\small
\label{eq:dimension}
D = dim(\overrightarrow{V_{F}}) = dim(\overrightarrow{V_{JI}}) = \displaystyle\prod_{k=1}^{C} G_{k}
\end{equation} 

where $\overrightarrow{V_{F}}$ and $\overrightarrow{V_{JI}}$ are the final and joint ideal vectors, $G_k$ is the number of groups in data characteristic k and C is the total number of data characteristics.
\subsection{Joint Ideal Percentage} \label{lable:jointideal}
In this section, we introduce Algorithm \ref{alg:jointideal} that calculates the \emph{"joint ideal percentage"} for each stratum. 
In this algorithm, $I_{g}^{c}$ represents the given ideal percentage for group g in characteristic c (e.g. 52\%\ female) and $JI_h$ is the joint ideal percentage in stratum h that is computed by multiplying the ideal percentages of all the characteristic groups involved in stratum h. In this paper, we assume that the data characteristics are independent. 

\begin{algorithm}
\small
\caption{Calculate joint ideal percentage for stratum h}
\begin{algorithmic}
\label{alg:jointideal}
\STATE $number\;of\;characteristics \leftarrow C $
\STATE $number\;of\;groups\;in\;characteristic\;c_k \leftarrow G_{k} $
\STATE $number\;of\;strata \leftarrow D $
\STATE $h \in \{1,\dots,D\}$
\STATE $h \leftarrow 1 $
\FOR{$i \in \{1,\dots,G_1\}$} 
 \FOR{$j \in \{1,\dots,G_2\}$} 
    \STATE $\;\;\vdots \notag$
    \FOR{$n \in \{1,\dots,G_C\}$} 
\STATE $JI_{h} = I_{g_i}^{c_1} \times I_{g_j}^{c_2} \times ... \times I_{g_n}^{c_C} $
\STATE $h \leftarrow h + 1 $
     \ENDFOR
    \STATE $\;\;\vdots \notag$
    \ENDFOR
\ENDFOR
\end{algorithmic}
\end{algorithm}
  
\subsection{Baseline Methods}\label{lab:baselines}
In this section, we describe the three different baseline methods that we compare with our proposed method in the experimental evaluation.\\
\noindent\textbf{Stratified Random Sampling}: Stratified Random Sampling (SRS) is a probability sampling technique in which the researcher partitions the entire population into strata, then randomly selects the final subjects from the strata\cite{taherdoost2016sampling,etikan2017sampling}.  
There are two types of stratified sampling techniques: Proportionate and Disproportionate. In \emph{Proportionate Stratified Random Sampling (PSRS)} the sample size used per stratum is proportional to the size of that stratum. The sample size in each stratum $n_{h}$ is determined by the following equation: 
\begin{equation} \small \label{eq:proportinate} n_{h}=\frac{N_{h}}{N}*n\end{equation}     
where $N_{h}$ is the population size for stratum h, N is the total population size and n is the total sample size\cite{wiki2,haque2010sampling}. If we remove n from this equation and multiply the remaining fraction by 100, we get the final percentage in each stratum obtained from this baseline.
With the disproportionate stratification, there is an intentional over-sampling of certain strata. In practice, disproportionate stratified sampling will only minimize sampling variance if the sampling fraction of each stratum is proportional to the standard deviation within that stratum. In other words, we should sample more from the more variable strata. This technique is called \emph{Optimum Stratified Random Sampling (OSRS)}\cite{wiki2}. We used a well-known method of optimum allocation called \emph{Neyman Allocation}\cite{neyman1992two,lavrakas2008encyclopedia} in our experiments with the following formula:
\begin{equation} \small \label{eq:neyman} n_{h}=\frac{N_{h}*\sigma_{h}}{\sum_{i=1}^{H}{N_{i}*\sigma_{i}}}*n
\end{equation}
     
where $N_{h}$ is the population size for stratum h, $\sigma_{h}$ is the standard deviation of stratum h and n is the total sample size. In practice, we do not know $\sigma_{h}$ in advance, so we randomly draw a set of samples and use them to derive an estimate of $\sigma_{h}$ for each stratum\cite{mathew2013efficiency}.
    
The percentages of the final selection for this baseline can be computed by dividing the Neyman formula by n and multiplying it by 100. 

\noindent\textbf{Rank Aggregation}: \emph{Rank Aggregation (RA)} is the process of merging multiple rankings of a set of subjects into a single ranking. There are two variations of rank aggregation: score-based and order-based. In the first category, the subjects in each ranking list are assigned scores and the rank aggregation function uses these scores to create a single ranking. In the second category, only the orders of the subjects are known and used by the rank aggregation function\cite{fox1994combination}.
However, if we have the scores we can either use score-based algorithms (e.g. Fagin’s algorithm\cite{fagin2003optimal}) or we can convert the scores to the orders and then apply order-based algorithms. According to our experiments, order-based algorithms return better results for our problem. 

In order to score the subjects, first, we define a weight function (equation \ref{eq:weight}) which is the ratio of the ideal percentage to the initial percentage of the characteristic group that each subject belongs to. As a result, a list of weights corresponding to each data characteristic is created and can be converted to the orders (the higher the weight, the lower the order). The list of orders is then used as input to an order-based rank aggregation method which produces a single ranked list with minimum total disagreement among the input lists.
 \begin{equation} 
 \small
 \label{eq:weight} 
 W_{s}^c= \frac{I_{g}^c}{Init_{g}^c}
 \end{equation}    
In this function, $W_{s}^c$ is the weight of subject s for data characteristic c, $I_{g}^c$ is the given ideal percentage for group g in characteristic c, $init_{g}^c$ is the initial percentage of group g in characteristic c which is obtained based on the ratio of the number of subjects in group g to the total population size.

To measure the disagreement between the input lists, the rank aggregation method can use either the \emph{Kendall Tau Distance} or the \emph{Spearman Footrule Distance}. The aggregation obtained by optimizing the Kendall distance is called \emph{Kemeny optimal aggregation} which is NP-Hard and the aggregation obtained by optimizing the Footrule distance is called \emph{Footrule optimal aggregation} which is a polynomial-time algorithm\cite{dwork2001rank,sculley2007rank}. The second algorithm is more efficient and so was selected as our rank aggregation method.
After applying the Footrule optimal aggregation on the lists of the ordered subjects, we select the top-n (n = total sample size) subjects from the final ranked list and then compute the percentage of the final selection in each stratum.

\noindent\textbf{Weighted Random Sampling}: \emph{Random Sampling (RS)} without replacement is the selection of m distinct random subjects out of a population of size n. If the probability of selection of each subject is the same as others', the problem is called \emph{Uniform Random Sampling}. However, in our problem we want each subject to have a different probability to be selected (based on the stratum it belongs to) which is not the case for uniform random sampling. Instead, we should use a \emph{Weighted Random Sampling (WRS)} in which subjects are weighted and the probability of selecting each subject is defined by its relative weight\cite{efraimidis2006weighted}. If in the subject selection problem we only address one data characteristic, the weight of each subject can be defined by the function we proposed in equation \ref{eq:weight}. However, in most real applications we have to deal with several characteristics involved in each stratum so we need to define a \emph{"joint weight"} function for each stratum (equation \ref{eq:jointweight}). Since each subject belongs to only one stratum, the joint weight of each subject will be equal to the weight of the stratum that subject belongs to.
 \begin{equation} 
 \small
 \label{eq:jointweight} 
 JW_{s} = JW_{h} = \frac{JI_{h}}{JInit_{h}}
 \end{equation}     
In this equation, $JI_{h}$ is the joint ideal percentage for stratum h (obtained by algorithm \ref{alg:jointideal}) and $JInit_{h}$ is the joint initial percentage obtained by the "joint frequency" concept. In other words, $JInit_{h}$ is the number of subjects belongs to stratum h divided by the total population size.
     
After calculating the joint weight, the probability of each subject to be selected can be defined by the product of its weight and its uniform probability:
\begin{equation}
\small
\label{eq:probabilty} 
P(s)= JW_{s}*1/N 
\end{equation}  
where $JW_s$ is the joint weight assigned to subject s in stratum h and 1/N is the probability of selecting a subject randomly form a population with size N where all subjects have an equal opportunity to be selected.

Having the assigned probabilities, a random sampling function can select n subjects (n = total sample size) from the population which results in computing the final percentage for each stratum. 
\subsection{Baseline Method Variations}
In this section, we explain how our baseline methods, presented in the previous section, address the three problem variations (fixed, range-based, and generalized). For all of the baseline methods, the joint ideal percentage for each stratum is computed by algorithm \ref{alg:jointideal}. The baseline method variations are as follows:
\begin{itemize}
 
   \item \textbf{Fixed}: in this variation, given the fixed ideal percentages for all data characteristic groups, each baseline introduces a method that generates the distribution of the final selection. 

   \item \textbf{Range-based}: in this variation, some or all of the ideal percentages are provided in the form of ranges. For this purpose, each baseline needs to iterate over all the possible enumerations that can be generated from a given range (must sum up to 100\%). Each baseline with every ideal percentage enumeration returns a different joint ideal distribution, a different final distribution and thus a different cosine distance. So, after going through all the feasible enumerations, we should find the smallest cosine distance and pick the final distribution, that generates this distance.

   \item \textbf{Generalized}: in this variation, a set of constraints or objective functions need to be addressed in addition to the given ranges. However, considering a set of constraints or adding new objective functions to the problem adds more complexity to the baseline methods so they will not be able to handle this variation very well.
    
\end{itemize}

\section{Proposed Method} 
As mentioned earlier, our goal is to develop an approach that finds a final distribution that is as close as possible to the joint ideal distribution. The baseline methods we explained in the previous section introduce different ways to select required subjects from a population but do not have any suggestions for minimizing the distance between the final and the joint ideal percentage vectors. 
Furthermore, in range-based variations, baselines might end up with too many enumerations, even if there are only a few data characteristics or a few groups involved in the subject selection problem. For example, Suppose a researcher wants to consider only three characteristics namely age, gender, and race in a subject selection problem such that each data characteristic consists of three groups and each group is provided with a range of length five. As previously mentioned, in the range-based problems, our baseline methods have to generate all the possible enumerations for all characteristics. For our particular example, this means we will end up with at least $5.12\times10^{10}$ combinations to try out 
(if every time we increase by 0.25 point within a range). In reality, the number of characteristics or characteristic groups can be even higher than what is given in this motivating example.\\
Moreover, in some real applications, a set of constraints or objective functions need to be satisfied within the subject selection but they will add more complexity to our baseline methods and so the baselines will not be able to meet all the given constraints in a simple way.\\ 
Considering all the above issues, we came up with an optimization-based approach that finds an optimal solution for minimizing the cosine distance between the joint ideal and final distributions, in the presence of a set of constraints. This approach is quite faster and less expensive in comparison to the baseline methods, especially for the range-based and generalized variations. Our proposed method supports all the problem variations we introduced as follows: 
\begin{itemize}
\item \textbf{Fixed}:  
    in this variation of the optimization problem, the objective function is defined to minimize the cosine distance between the final percentage vector $\overrightarrow{V_{F}} = (F_{1}, F_{2}, ...\;, F_{D})$ and the joint ideal percentage vector $\overrightarrow{V_{JI}} = (JI_{1}, JI_{2}, ...\;, JI_{D})$.    
    In formulation \ref{eq:optfixed}, the cosine distance is defined based on equations \ref{eq:distance} and \ref{eq:cosine}, $F_h$ (for h=1,2,...,D) represents the components of $\overrightarrow{V_{F}}$ that are defined as variables and $JI_h$ (for h=1,2,...,D) represents the components of $\overrightarrow{V_{JI}}$ that are defined as constants.
    The objective function needs to be optimized subject to a constraint on the sum of the final percentages (the sum of final percentages is equal to 1, not 100 because we converted each percentage to a fraction) and a boundary where the final number of selected subjects in each stratum ($F_h * n$: n is the total sample size) must be smaller than the initial number of subjects ($init_{h}$) in that stratum.
    \smallskip \begin{equation}
\small
\setlength\abovedisplayskip{0pt}
\setlength\belowdisplayskip{0pt}
\label{eq:optfixed}
\begin{array}{rrclcl}
\displaystyle \min_{F_h} & \multicolumn{3}{l}
{CosineDistance(\overrightarrow{V_{F}},\overrightarrow{V_{JI}})} \\
\textrm{s.t.} &\displaystyle \sum_{h=1}^{D} F_h & = & 1 \\
& F_h & \leq & init_{h}/n \\ 
& F_h & \geq & 0 & & h = 1,2,...,D\\
\end{array}
\end{equation}
 \\
    
\item \textbf{Range-based}: 
    since in this variation, some or all the provided ideal percentages are range-based, we consider the range-based ideal percentages as variables and the fixed ones as constants. However, in this optimization problem, we assumed that all of the ideal percentages are range-based so they are all determined as variables and their given ranges (e.g. $[a_k,b_k]$) are specified as the boundaries of these variables (one boundary per group). There is also one constraint defined per each characteristic in which the ideal percentages of its groups must add up to 1. The remaining of this formulation including the variables for the final percentages, its constraint, and boundaries is similar to the fixed variation in formulation \ref{eq:optfixed}.

    Formulation \ref{eq:optrange} represents the optimization problem for a range-based variation with C characteristics including $\{c_1, c_2, ..., c_C\}$ whose number of groups are $\{G_1, G_2, ..., G_C\}$. In this optimization problem, the objective function is constructed using equations \ref{eq:distance} and \ref{eq:cosine}, $F_h$ is the final percentage for stratum h, $JI_h$ is the joint ideal percentage for stratum h that is formulated in algorithm \ref{alg:jointideal} where $I_{g}^{c}$ is the ideal percentage for group g in characteristic c, $init_{h}$ is the initial number of subjects in stratum h and n is the total sample size. The optimal solution to this optimization model consists of the components of $\overrightarrow{V_{F}}$ and $\overrightarrow{V_{IJ}}$ that are real numbers between 0 and 1 and the optimal objective function value to it, is the cosine distance between the two vectors.
   \bigskip 
\begin{equation}
    \small
    \setlength\abovedisplayskip{0pt}
    \setlength\belowdisplayskip{0pt}
    \label{eq:optrange}
    \tag{10}
    \begin{array}{rrclcl}
    \displaystyle \min_{F_h,JI_h} 
    & \multicolumn{3}{l}{{CosineDistance(\overrightarrow{V_{F}},\overrightarrow{V_{JI}})}} \\
    \textrm{s.t.} &\displaystyle \sum_{h=1}^{D} F_h & = & 1 \\
    &\displaystyle \sum_{k=1}^{G_1} I_{g_{k}}^{c_1} & = & 1 \\
    &\displaystyle \sum_{l=1}^{G_2} I_{g_{l}}^{c_2} & = & 1 \\
    & \;\;\vdots \notag & &\\
    &\displaystyle \sum_{m=1}^{G_C} I_{g_{m}}^{c_C} & = & 1 \\
    &I_{g_{k}}^{c_1} & \in & [a_k,b_k]\quad\;\;\;a_k,b_k\in[0,1]\;,k=1,2,..., G_1\\
    &I_{g_{l}}^{c_2} & \in & [a^{\prime}_l,b^{\prime}_l]\quad\quad\;a^{\prime}_l, b^{\prime}_l \in [0,1]\;\;,l=1,2,..., G_2\\
    & \;\;\vdots \notag & &\\
    &I_{g_{m}}^{c_C} & \in & [a^{"}_m,b^{"}_m]\quad\; a^{"}_m, b^{"}_m \in [0,1]\;,m=1,2,..., G_C\\
    & F_h & \leq & init_{h}/n \quad\\ 
    & F_h & \geq & 0\quad\quad\quad\quad\;  h = 1,2,...,D \\
    \end{array}
\end{equation}

\item\textbf{Generalized}: 
    we can simply generalize the range-based variation in formulation \ref{eq:optrange} by adding the required equality and inequality constraints on the specific characteristic groups. 
     For example, if the required constraint is to select twice as many female subjects as male subjects, we should add $I_{{female}}^{gender} - 2*I_{{male}}^{gender} = 0$ as a constraint to the specific version of formulation \ref{eq:optrange}. In terms of adding new objectives (e.g. motivating example \#2), the existing problem should address more than one objective function simultaneously and transform to a multi-objective optimization problem.\\
    
\end{itemize}
\begin{table*}[ht!]
\small
\centering
\caption{datasets and their configurations}
\vspace{0.5cm}
\label{tab:datainfo}
\begin{tabular}{l|c|c|c|c|l}
\multicolumn{1}{c|}{\textbf{datasets}} & \textbf{\begin{tabular}[c]{@{}c@{}}total\\ population size\end{tabular}} & \textbf{\begin{tabular}[c]{@{}c@{}}total\\ sample size\end{tabular}} & \textbf{\begin{tabular}[c]{@{}c@{}}fixed ideal \\ distribution\end{tabular}} & \textbf{\begin{tabular}[c]{@{}c@{}}number of \\ characteristics\end{tabular}} & \multicolumn{1}{c}{\textbf{characteristics}} \\ \hline
Pittsburgh survey data & 889 & 100 & census (3 cities) & 2 & gender, age \\
Kaggle Titanic data & 714 & 100 & custom & 3 & gender, age, ticket class \\
Kaggle loan data & 500 & 100 & uniform & 3 & gender, age, education \\
Github loan data & 598 & 100 & uniform & 4 & gender, marital status, education, area  
\end{tabular}
\end{table*}
\noindent \textbf{Complexity:} \\
If we have k data characteristics and each characteristics has n groups then the number of variables in the optimization-based approach (all variations) will be $O(n^k)$ and the number of constraints will be $O(1)$. In terms of time complexity, running the proposed optimization-based approach (for all experiments) on a modern laptop takes less than a second while running the baseline methods takes a few seconds for the fixed variation and a few minutes for the range-based variations. 
\section{Experiments} \label{sec:experiments}
\begin{figure}[!hb]
  \centering
  \includegraphics[width=1\linewidth]{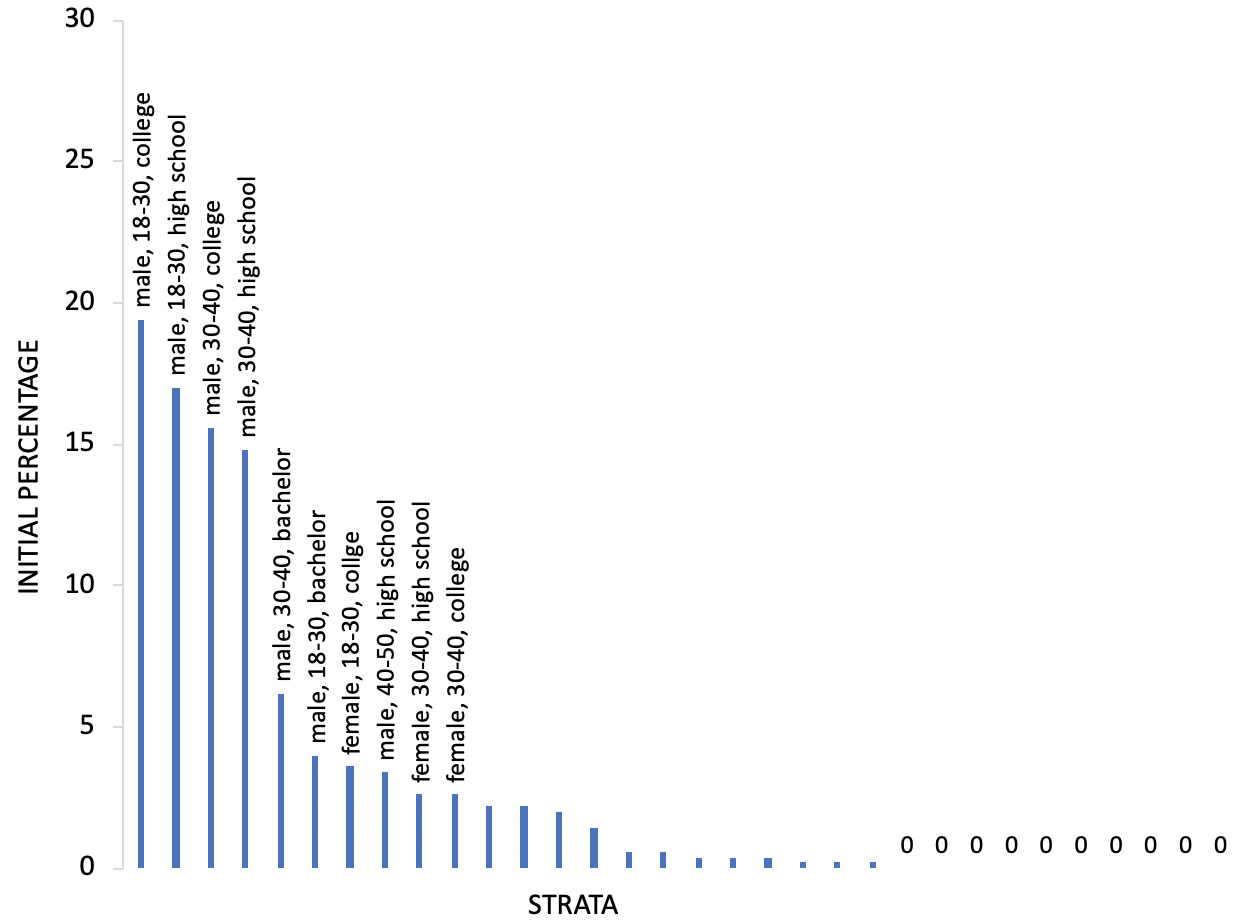}
  \caption{joint initial distribution for Kaggle Loan Data}
  \label{fig:kloan_initial}
\end{figure}

For the experimental evaluation of this work, our code-base was
written in Python 3. We used cosine distance as the evaluation metric and four different datasets to evaluate the baseline methods and our proposed approach. In each experiment, we compare the optimal objective function value achieved by the optimization-based method with the cosine distance obtained from the baselines.
It should be pointed out that we examined and compared several of the available solvers that were applicable to our optimization problems (constrained nonlinear programming with boundaries) including gradient-based (e.g., SLSQP) and gradient-free (e.g., COBYLA and Ant Colony). Among these algorithms, SLSQP\cite{kraft1988software} (supported by Python scipy.optimize.minimize package) was the fastest solver with the most desirable results. 
\subsection{Datasets (Table \ref{tab:datainfo} , Figure \ref{fig:kloan_initial})} \label{lab:data} 
We used four different datasets for evaluating the baselines and the optimization-based method. Pittsburgh survey data was collected through an online survey of transportation preferences of adults from a diverse population (in terms of age, income, etc) in the broader Pittsburgh area. Kaggle Loan data\cite{kaggleloan} and Kaggle Titanic data\cite{kaggletitanic} are open datasets and received from the Kaggle website, and Github Loan data was collected from Github. More information about these datasets and the configurations is presented in Table \ref{tab:datainfo}. \\
Furthermore, Figure \ref{fig:kloan_initial} illustrates the joint initial distribution for all the strata in Kaggle loan data (only the top 10 strata with the highest initial percentages are labeled). As one can see in this figure, the data is biased and not representative of the general population, since it is male-dominated.
\begin{figure*}[!h]
\setlength{\tabcolsep}{10pt}
\begin{minipage}[b]{0.5\linewidth}\centering

\includegraphics[width=\linewidth]{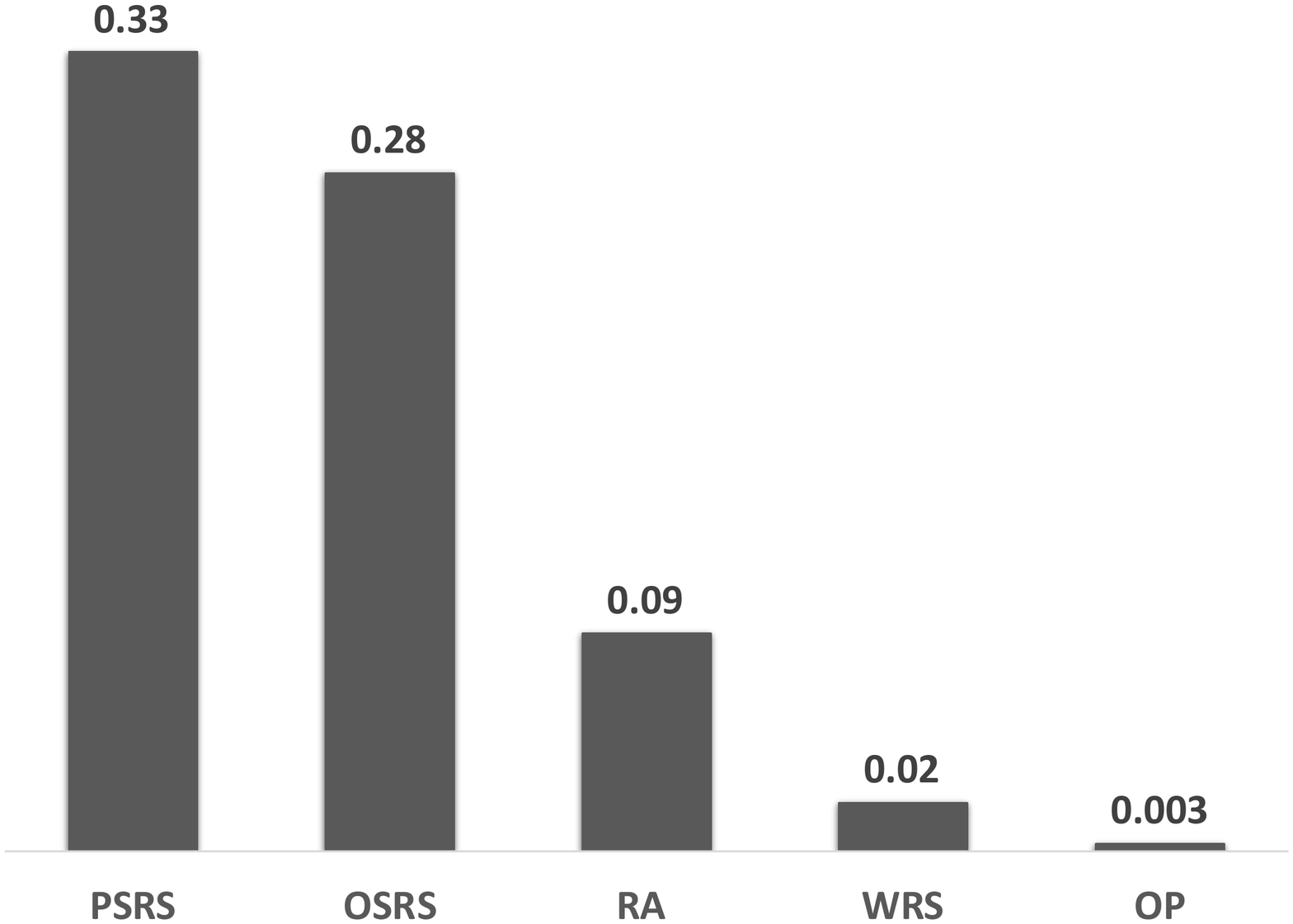}
\caption{comparison of cosine distance obtained by baselines (PSRS, OSRS, RA \&\ WRS) and optimization-based (OP) method using Kaggle Titanic data with only one characteristic}
\label{fig:oneVar_results}
\end{minipage}
\hspace{0.5cm}
\begin{minipage}[b]{0.5\linewidth}
\includegraphics[width=\linewidth]{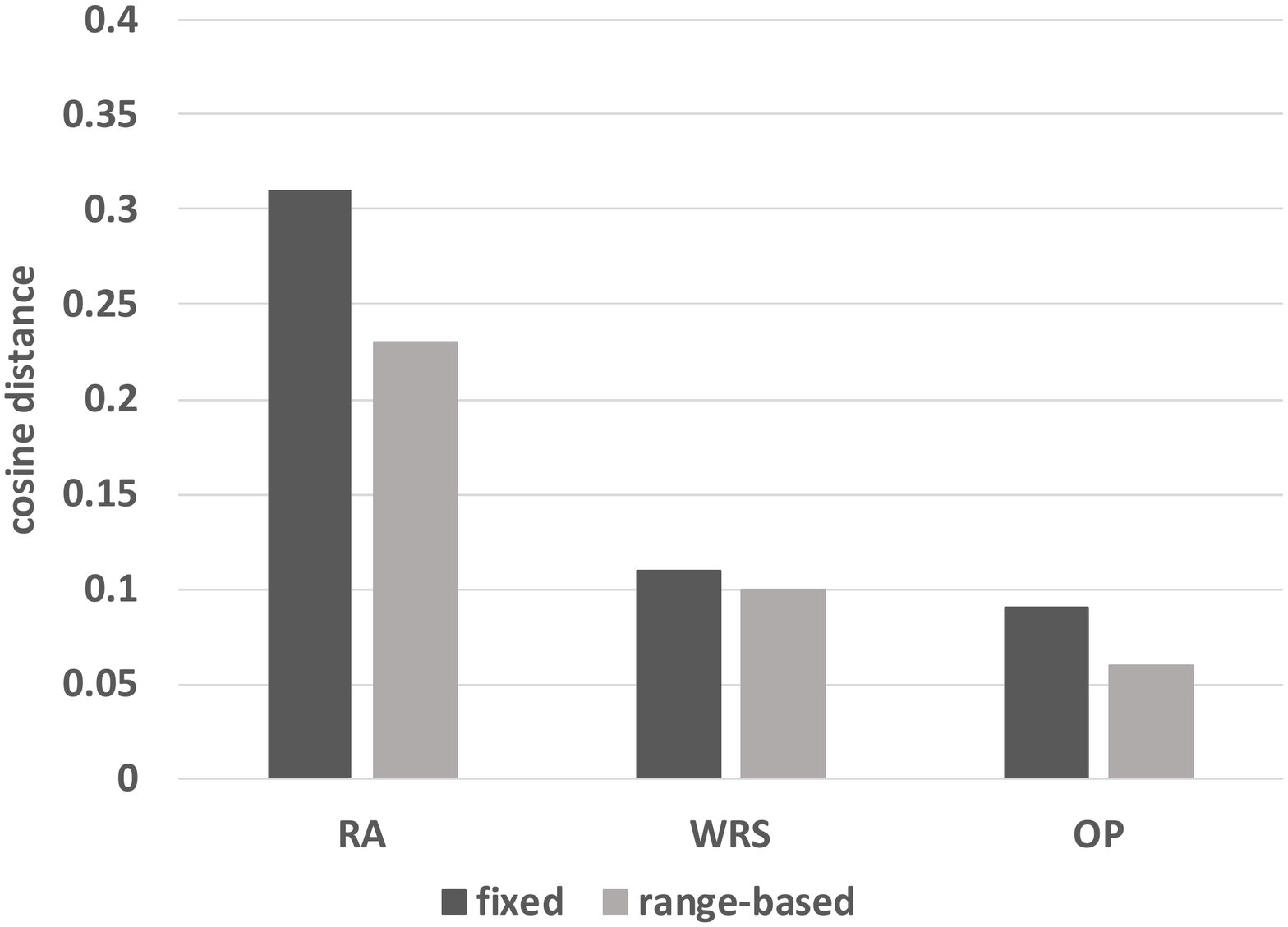}
\caption{comparison of cosine distance for fixed and range-based variations obtained by baselines and optimization-based (OP) method for survey data with ideal distribution from Pittsburgh census.}
\label{fig:pitt_fixed_range}

\end{minipage}
\end{figure*}
\subsection{Baselines vs Optimization-based Method - One Characteristic (Figure~\ref{fig:oneVar_results})}
\begin{figure*}[h!] 
\centering %
  \begin{subfigure}[b]{0.30\textwidth}
    \includegraphics[width=\linewidth]{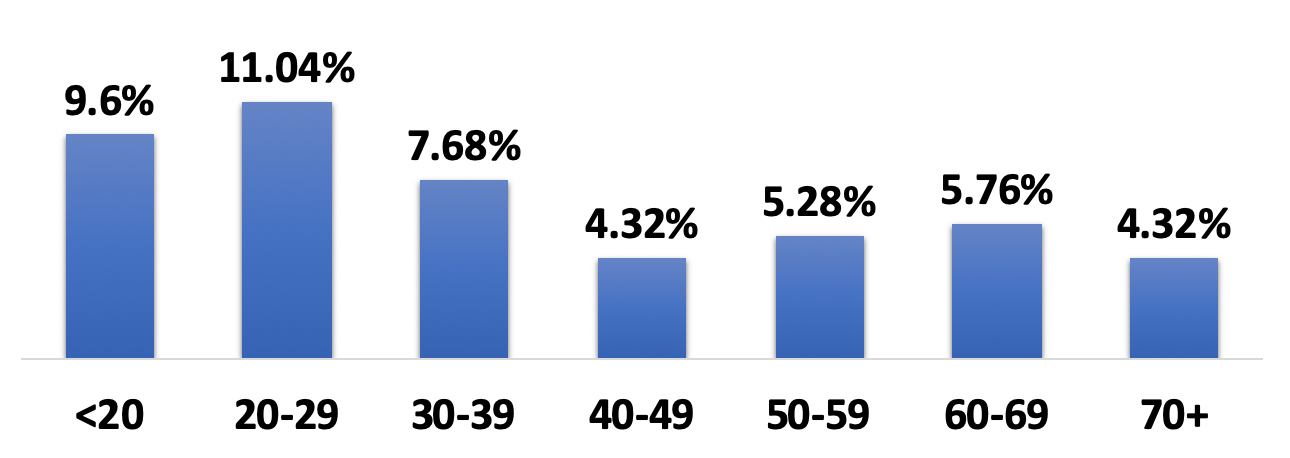}
    \caption{joint ideal - male} 
    \label{fig:pitt_ideal_male} 
  \end{subfigure}\hfil %
  \begin{subfigure}[b]{0.30\textwidth}
    \includegraphics[width=\linewidth]{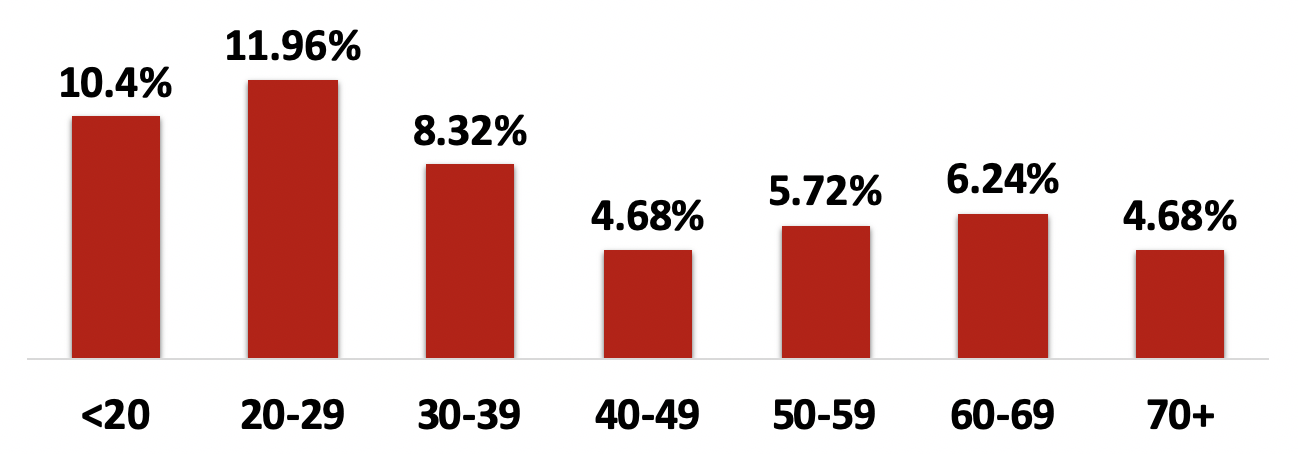}
    \caption{joint ideal - female} 
    \label{fig:pitt_ideal_female} 
 \end{subfigure}\hfil %
  \begin{subfigure}[b]{0.30\textwidth}
    \includegraphics[width=0.45\linewidth]{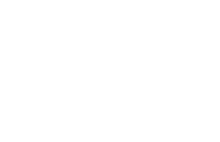}
  \end{subfigure}
  
  \bigskip
  \begin{subfigure}[b]{0.30\textwidth}
    \includegraphics[width=\linewidth]{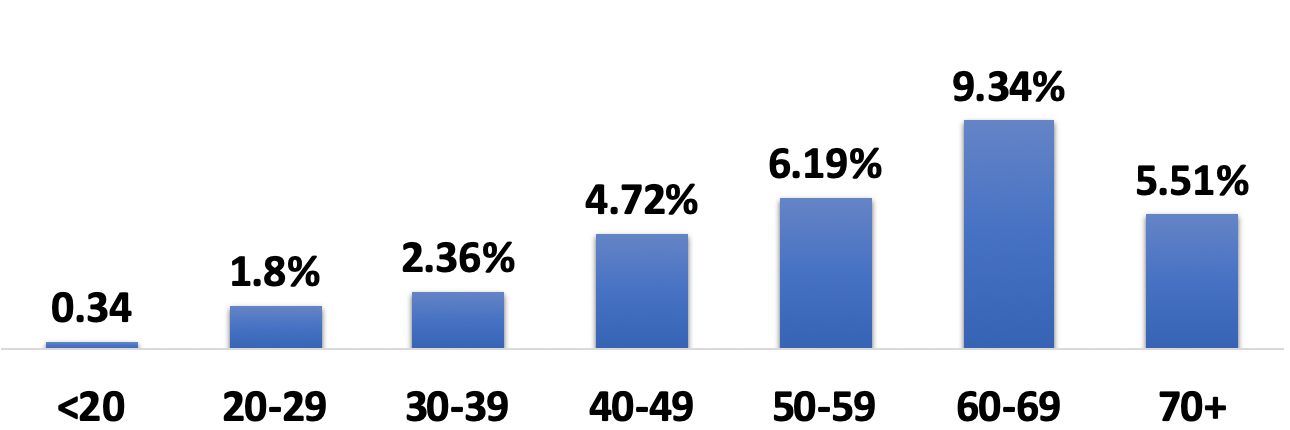}
    \caption{joint initial - male} 
    \label{fig:pitt_initial_male} 
  \end{subfigure}\hfil %
  \begin{subfigure}[b]{0.30\textwidth}
    \includegraphics[width=\linewidth]{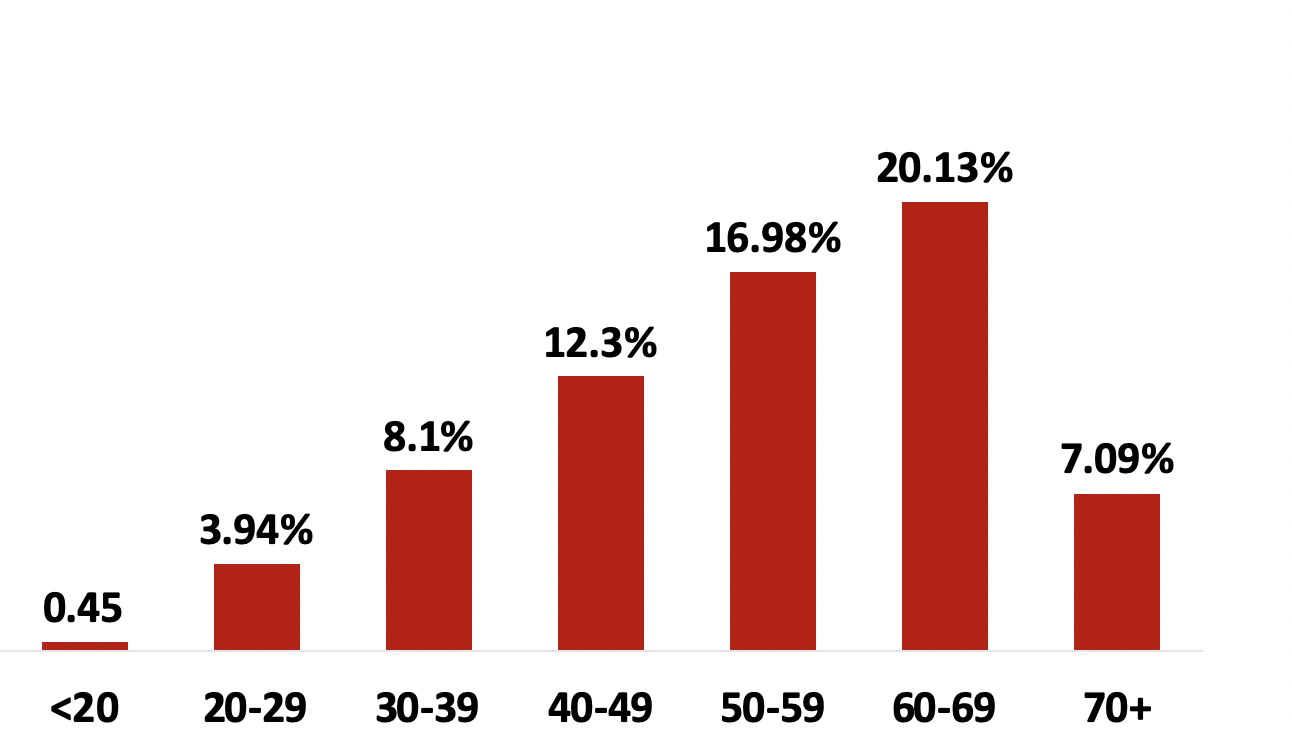}
    \caption{joint initial - female} 
    \label{fig:pitt_initial_female} 
  \end{subfigure}\hfil %
  \begin{subfigure}[b]{0.30\textwidth}
  \centering
    \includegraphics[width=0.45\linewidth]{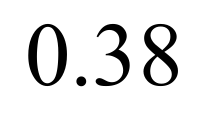}
    \caption{cosine distance} 
    \label{fig:pitt_initial_cosine} 
  \end{subfigure}
  
  \bigskip
  \begin{subfigure}[b]{0.30\textwidth}
    \includegraphics[width=\linewidth]{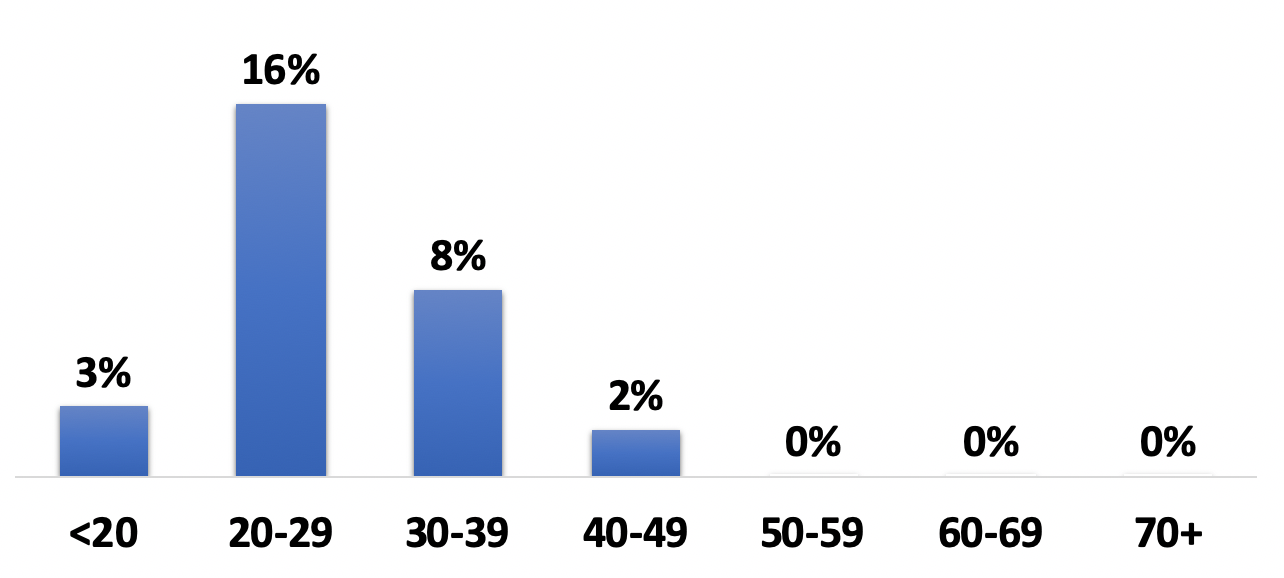}
    \caption{final dist. (RA) - male} 
    \label{fig:pitt_RA_male} 
  \end{subfigure}\hfil %
  \begin{subfigure}[b]{0.30\textwidth}
    \includegraphics[width=\linewidth]{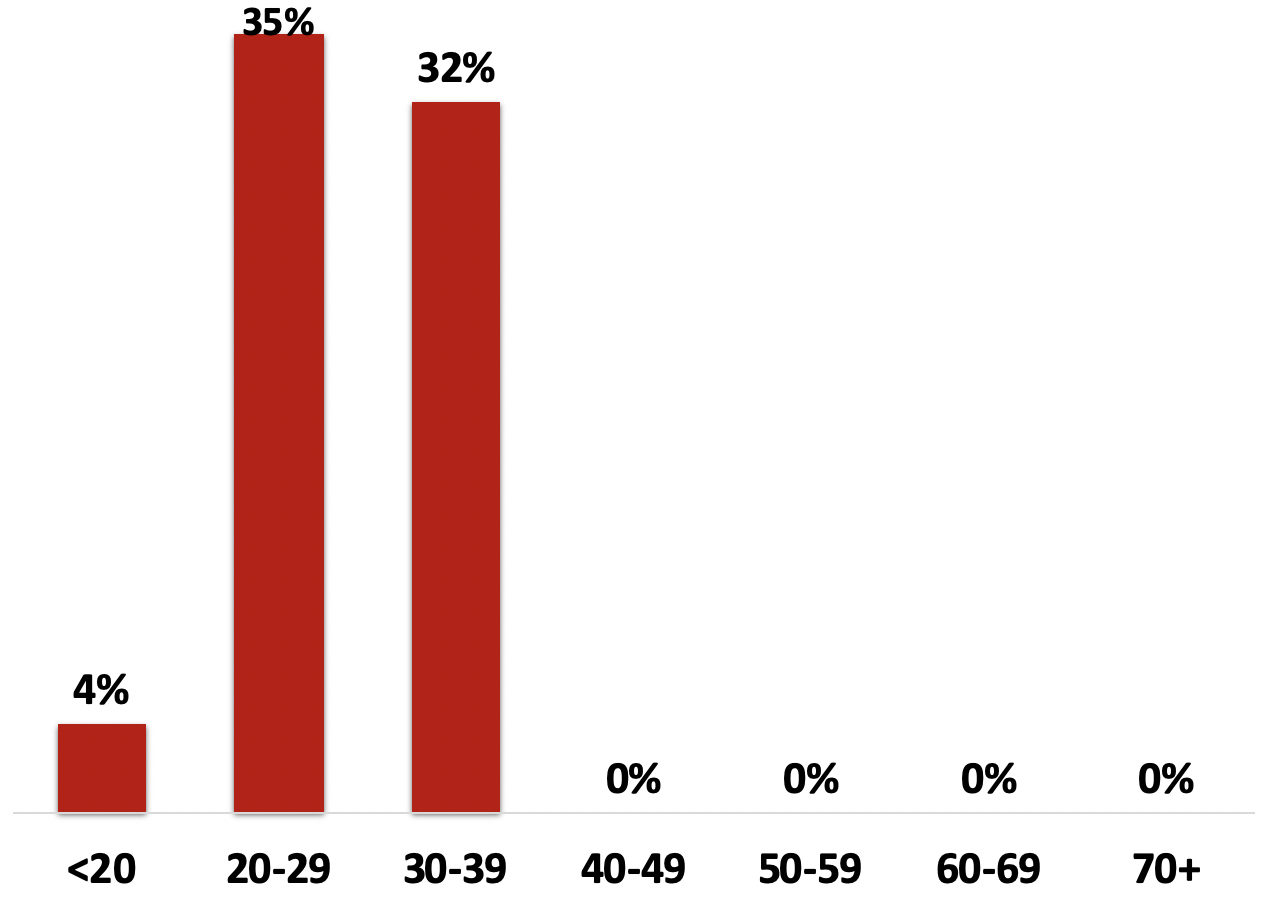}
    \caption{final dist. (RA) - female} 
    \label{fig:pitt_RA_female} 
  \end{subfigure}\hfil %
  \begin{subfigure}[b]{0.30\textwidth}
  \centering
    \includegraphics[width=0.45\linewidth]{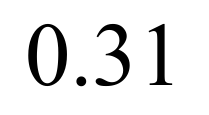}
    \caption{cosine distance} 
    \label{fig:pitt_RA_cosine} 
  \end{subfigure}
  
  \bigskip
  \begin{subfigure}[b]{0.30\textwidth}
    \includegraphics[width=\linewidth]{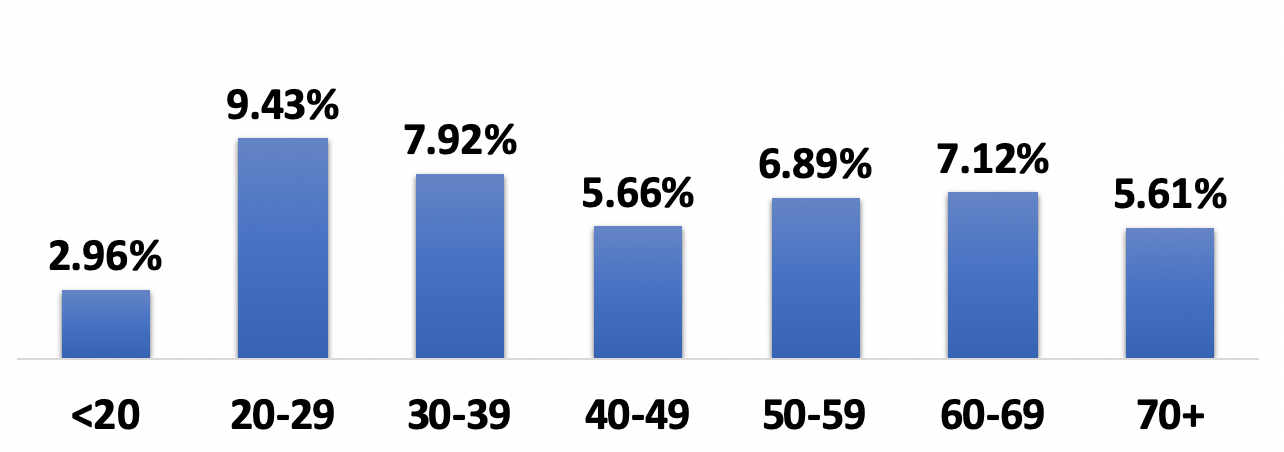}
    \caption{final dist. (WRS) - male} 
    \label{fig:pitt_WRS_male} 
  \end{subfigure}\hfil %
  \begin{subfigure}[b]{0.30\textwidth}
    \includegraphics[width=\linewidth]{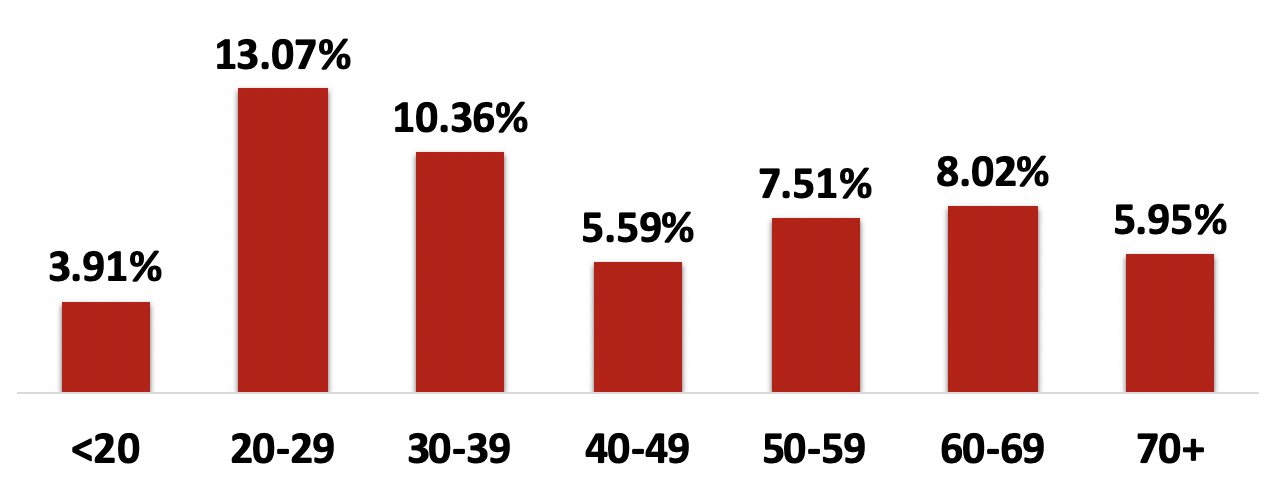}
    \caption{final dist. (WRS) - female} 
    \label{fig:pitt_WRS_female} 
  \end{subfigure}\hfil %
  \begin{subfigure}[b]{0.30\textwidth}
  \centering
    \includegraphics[width=0.45\linewidth]{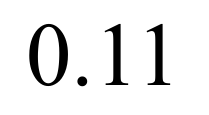}
    \caption{cosine distance} 
    \label{fig:pitt_WRS_cosine} 
  \end{subfigure}
  
  \bigskip
  \begin{subfigure}[b]{0.30\textwidth}
    \includegraphics[width=\linewidth]{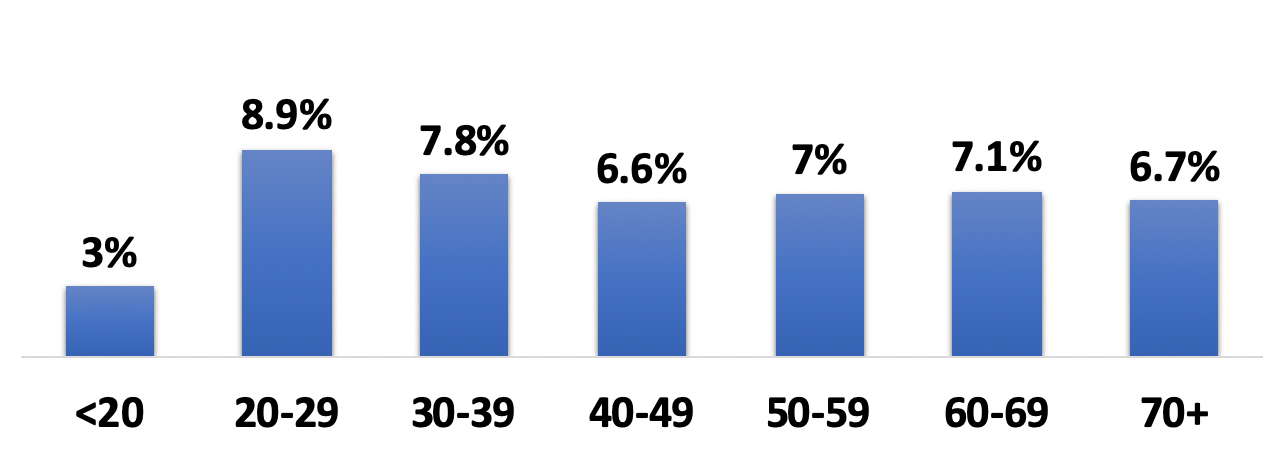}
    \caption{final dist. (OP) - male} 
    \label{fig:pitt_OP_male} 
  \end{subfigure}\hfil %
  \begin{subfigure}[b]{0.30\textwidth}
    \includegraphics[width=\linewidth]{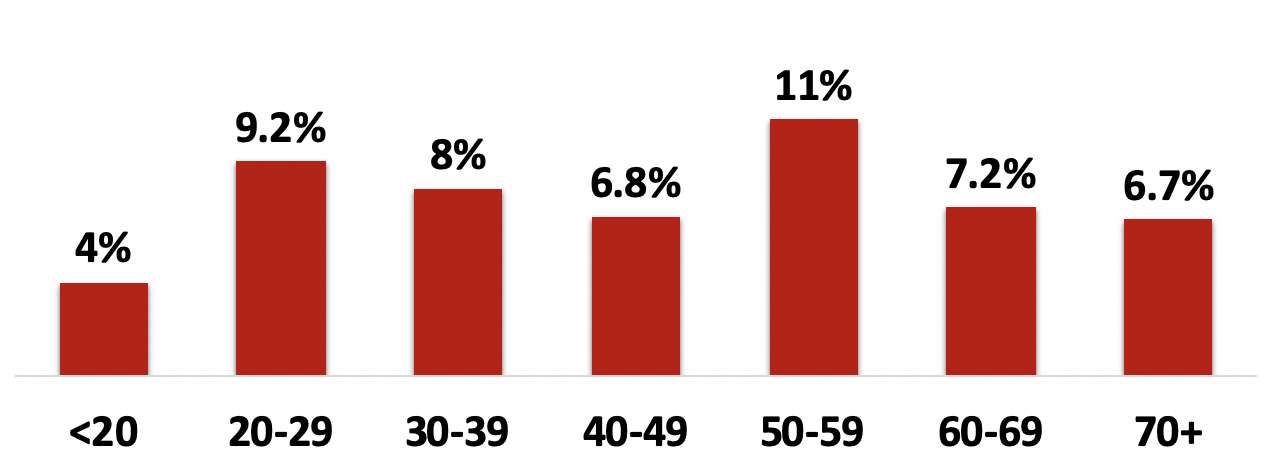}
    \caption{final dist. (OP) - female} 
    \label{fig:pitt_OP_female} 
  \end{subfigure}\hfil %
  \begin{subfigure}[b]{0.30\textwidth}
  \centering
    \includegraphics[width=0.45\linewidth]{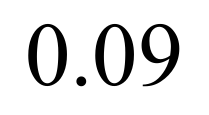}
    \caption{cosine distance} 
    \label{fig:pitt_OP_cosine} 
  \end{subfigure}
  
\caption{joint ideal, joint initial, final dist. for age groups and cosine distance obtained by the baseline methods: RA (Rank Aggregation), WRS (Weighted Random Sampling) and OP (Optimization-based) applied on Pittsburgh survey data}
\label{fig:pitt_panel}   
\end{figure*}

{\noindent\bf Setup:} in this experiment, we evaluate our baselines and the optimization-based method using one of our datasets called "Kaggle Titanic", considering "age" as the data characteristic, along with a custom fixed ideal distribution given for the age groups. 
Since we only have one characteristic, the age groups themselves can be considered as the strata. The age groups include \{-10,
10-20, 20-30, 30-40, 40-50, 50-60, 60+\} and their assigned ideal distribution is \{40\%, 30\%, 20\%, 10\%, 0\%, 0\%, 0\%\} where young people have higher priority to be selected.\\ 
{\noindent\bf Results:} as one can see in Figure \ref{fig:oneVar_results}, the cosine distances obtained from the Proportionate Stratified Random Sampling (PSRS) and the Optimal Stratified Random Sampling (OSRS) are quite higher than the other methods'. We exclude stratified sampling baselines from the next experiments due to this big difference. According to this figure, the proposed method with the optimization-based (OP) approach improved the cosine distance by 97\% and 85\% with respect to to the RA and WRS methods.\\
{\noindent\bf Take-away:} the optimization-based method performed better (up to 97\%) than the baselines in the experiment that was conducted with one data characteristic and a custom fixed ideal distribution. As a result, the stratified random sampling methods were not considered in the following experiments due to their unacceptable performance.
\subsection{Baselines vs Optimization-based Method - Multi-characteristics (Figures~\ref{fig:pitt_fixed_range}, \ref{fig:pitt_panel} and Tables~\ref{tab:allsurveycomp}, \ref{tab:alldatacomp})}
\noindent\textbf{1- Experiments with Pittsburgh Survey Data (Figures~\ref{fig:pitt_fixed_range}, \ref{fig:pitt_panel}):} 

{\noindent\bf Setup:} this experiment is done using the Pittsburgh survey data, focusing on two data characteristics including: age and gender. In the fixed variation, the ideal distributions for age and gender groups are collected from the Pittsburgh census reporter\cite{censusreporter}. In the range-based variation, the ideal distribution for age groups are fixed and from census, but the gender groups are provided with a range-based ideal distribution. 

{\noindent\bf Results:} Figure \ref{fig:pitt_fixed_range} shows the comparison between the two baselines (RA and WRS) and the optimization-based (OP) method for the two variations of the problem: fixed and range-based. The optimization-based method improves the distance by 18\% and 40\% in comparison to WRS, and by 71\% and 74\% in comparison to RA, for the fixed and range-based variations respectively. On the other hand, the experiments in Figure \ref{fig:pitt_panel} were conducted for the fixed variation only. The x-axis represents the age groups and the y-axis (the scale is 0\% to 35\%) shows the percentage of male/female in each gender group. As indicated, Figures \ref{fig:pitt_ideal_male} and \ref{fig:pitt_ideal_female} show the joint ideal distributions for the age groups separated by male and female. These values were computed by algorithm \ref{alg:jointideal} using the age and gender distributions from the Pittsburgh census. 
Figures \ref{fig:pitt_initial_male} and \ref{fig:pitt_initial_female} illustrate the joint initial distributions extracted from the survey data whose distance from the joint ideal distribution (shown in Figures \ref{fig:pitt_ideal_male} and \ref{fig:pitt_ideal_female}) is 0.38. The total sum of the joint initial percentages in the two figures is not equal to 100\%. The reason is that there are other gender groups (neither male nor female) in the Pittsburgh survey data which are excluded from this distribution.
In the next three rows of the Figure \ref{fig:pitt_panel} including \ref{fig:pitt_RA_male} to \ref{fig:pitt_OP_female}, we represent the final distributions as the outputs of our two baseline methods (RA and WRS) and the proposed method (OP). When the cosine distance between the final distributions and the joint ideal distribution becomes smaller, the final distributions become more similar to the joint ideal distribution. It is worth mentioning that we have about 20\% improvement (from 0.11 to 0.09) in the distance when using the optimization-based method compared to the WRS method and about 70\% improvement (from 0.31 to 0.09) in comparison with the RA method. 
\begin{table*}[ht!]
\small
\setlength{\tabcolsep}{8pt}
\centering
\caption{cosine distance obtained by baselines and optimization-based (OP) method for Pittsburgh survey data with three census dist.}
\label{tab:allsurveycomp}
\begin{tabular}{c|c|c|c|l|c|c|c|l|c|c|c|}
\cline{2-4} \cline{6-8} \cline{10-12}
 & \multicolumn{3}{c|}{\textbf{Pittsburgh Census}} &  & \multicolumn{3}{c|}{\textbf{New York Census}} &  & \multicolumn{3}{c|}{\textbf{College Park Census}} \\ \cline{2-4} \cline{6-8} \cline{10-12} 
\textbf{} & \textit{\textbf{RA}} & \textit{\textbf{WRS}} & \textit{\textbf{OP}} &  & \textit{\textbf{RA}} & \textit{\textbf{WRS}} & \textit{\textbf{OP}} &  & \textit{\textbf{RA}} & \textit{\textbf{WRS}} & \textit{\textbf{OP}} \\ \cline{1-4} \cline{6-8} \cline{10-12} 
\multicolumn{1}{|c|}{\textbf{Fixed}} & 0.31 & 0.11 & \textbf{0.09} &  & 0.42 & 0.14 & \textbf{0.10} &  & 0.38 & 0.23 & \textbf{0.21} \\ \cline{1-4} \cline{6-8} \cline{10-12} 
\multicolumn{1}{|c|}{\textbf{Range-based}} & 0.23 & 0.10 & \textbf{0.06} &  & 0.30 & 0.13 & \textbf{0.10} &  & 0.27 & 0.21 & \textbf{0.19} \\ \cline{1-4} \cline{6-8} \cline{10-12} 
\end{tabular}
\end{table*}
\begin{table*}[ht!]
\small
\setlength{\tabcolsep}{8pt}
\caption{cosine distance obtained by baselines and optimization-based (OP) method for three different datasets}
\label{tab:alldatacomp}
\centering
\begin{tabular}{c|c|c|c|l|c|c|c|l|c|c|c|}
\cline{2-4} \cline{6-8} \cline{10-12}
 & \multicolumn{3}{c|}{\textbf{Kaggle Loan Data}} &  & \multicolumn{3}{c|}{\textbf{Github Loan Data}} &  & \multicolumn{3}{c|}{\textbf{Kaggle Titanic Data}} \\ \cline{2-4} \cline{6-8} \cline{10-12} 
 & \textit{\textbf{RA}} & \textit{\textbf{WRS}} & \textit{\textbf{OP}} &  & \textit{\textbf{RA}} & \textit{\textbf{WRS}} & \textit{\textbf{OP}} &  & \textit{\textbf{RA}} & \textit{\textbf{WRS}} & \textit{\textbf{OP}} \\ \cline{1-4} \cline{6-8} \cline{10-12} 
\multicolumn{1}{|c|}{\textbf{Fixed}} & 0.35 & 0.14 & \textbf{0.13} &  & 0.46 & 0.09 & \textbf{0.05} &  & 0.47 & 0.25 & \textbf{0.22} \\ \cline{1-4} \cline{6-8} \cline{10-12} 
\multicolumn{1}{|c|}{\textbf{Range-based}} & 0.32 & 0.13 & \textbf{0.09} &  & 0.25 & 0.08 & \textbf{0.02} &  & 0.42 & 0.24 & \textbf{0.18} \\ \cline{1-4} \cline{6-8} \cline{10-12} 
\end{tabular}
\end{table*}



{\noindent\bf Take-away:} our proposed method outperformed the baseline methods (by up to 70\%) in the experiments conducted with the Pittsburgh survey data, with both fixed and range-based variations.\\

\noindent\textbf{2- Experiments with All Datasets (Tables~\ref{tab:allsurveycomp}, \ref{tab:alldatacomp}):} \\
{\noindent\bf Setup:} these experiments are done using all the datasets introduced in Table \ref{tab:datainfo} considering the outlined configurations. For the fixed variations, the ideal distributions are census-based, uniform (all groups in a data characteristic have the same ideal percentage) or custom (defined by us). For the range-based variations, we set the ideal distribution of the gender groups to be range-based and for the other characteristic groups to be fixed.\\
{\noindent\bf Results:} Table \ref{tab:allsurveycomp} reports the cosine distances obtained from the baselines and the proposed method using the Pittsburgh survey data with fixed and ranged-based variations. For the fixed variation, we used three different census distributions from three different cities in the US including Pittsburgh, New York, NY and College Park, MD\cite{censusreporter}. As can be observed, our optimization-based method always outperforms its competitors for both variations. In particular, our proposed method improves the cosine distance up to 40\% compared to WRS and up to 76\% compared to RA. 
Table \ref{tab:alldatacomp} represents the comparison between cosine distance computed by the baseline methods and our proposed approach with three more datasets. The fixed ideal distributions that we used for these experiments are either uniform or customized. As numbers in the table indicate, the optimization-based method performs better than the baselines and improves the cosine distance up to 75\% compared to WRS and up to 90\% compared to RA. 

{\noindent\bf Take-away:} our proposed method outperformed the baseline methods (by up to 90\%) for the experiments conducted with all the datasets for both fixed and range-based variations over a different number of data characteristics.
\subsection{Discussion}
\begin{table}[hb!]
\centering
\small
\caption{a summary of comparison between the baselines and optimization-based (OP) method }
\label{tab:methods_comp}
\begin{tabular}{|c|c|c|c|c|} \hline
\diagbox[width=8em]{\textbf{Variations}}{\textbf{Methods}} & SRS & RA & WRS & OP \\ \hline
Fixed &  \checkmark & \checkmark & \checkmark & \checkmark \\ \hline
Range-based & ? & ? & ? & \checkmark \\ \hline
Generalized & \xmark & \xmark & \xmark & \checkmark \\ \hline
\end{tabular}
\end{table}
As previously stated, baselines and the proposed method can all handle the fixed variation of the problem very well. However, the baseline methods will face more complexity in the range-based variation especially in high dimensional space where the number of characteristics or characteristic groups is big. In the generalized variation, the complexity can get even worse or the baselines might not be able to handle the required constraints or objective functions at all. Table \ref{tab:methods_comp} summarizes our comparison between the baseline methods and the optimization-based method.\\
A different use case is one where the researcher running the study that does not have ideal percentages, but instead would like to \emph{select by example}. That would mean that the researcher could simply label the subjects as "desired" and "not-desired". Having the labeled data, we can count the number of desired subjects in each group for every data characteristic and divide it by the total number of the desired subjects. This way, we technically extract the hidden ideal percentage for each group and simply transform this case to a fixed variation.

\section{Conclusion}
In this paper, we proposed an optimization-based method for multi-characteristic subject selection from biased datasets. Our proposed method supports fixed and ranged-based variations along with any required constraints or objective functions on the characteristic groups.
We compared our method's performance with three baseline methods including stratified random sampling, rank aggregation, and weighted random sampling. Our experimental evaluation showed that our proposed method outperforms the baseline methods in all variations of the problem by up to 90\%. 




\begin {acks}
This work is part of the PittSmartLiving project which is supported by NSF award CNS-1739413.
\end{acks}


%
\bibliographystyle{ACM-Reference-Format}
\bibliography{biblio}
\end{document}